\documentclass{article}
\usepackage{iclr2026_conference,times}
\usepackage{wrapfig}
\usepackage{amsmath,amsfonts,bm}

\def\eqref#1{equation~\ref{#1}}

\def\1{\bm{1}}

\DeclareMathAlphabet{\mathsfit}{\encodingdefault}{\sfdefault}{m}{sl}
\SetMathAlphabet{\mathsfit}{bold}{\encodingdefault}{\sfdefault}{bx}{n}

\usepackage{booktabs}
\usepackage{graphicx}
\usepackage{xcolor}
\usepackage{hyperref}
\usepackage{url}
\usepackage[most]{tcolorbox}
\usepackage{listings}
\usepackage{xspace}
\usepackage{marvosym}
\usepackage{algorithm}
\usepackage{algorithmic}

\definecolor{hunyuanblue}{HTML}{0052D9}
\hypersetup{
    breaklinks=true,
    citecolor=hunyuanblue,
    colorlinks=true,
    linkcolor=hunyuanblue,
    urlcolor=hunyuanblue
}

\definecolor{PromptBack}{HTML}{F4FAFF}
\definecolor{PromptFrame}{HTML}{8EC5F4}
\definecolor{PromptTitle}{HTML}{1F5F99}

\lstdefinestyle{PromptBoxListing}{
  basicstyle=\small\ttfamily,
  columns=fullflexible,
  keepspaces=true,
  breaklines=true,
  breakatwhitespace=false,
  breakindent=0pt,
  postbreak={},
  showstringspaces=false,
  upquote=false,
  tabsize=2
}

\newtcblisting{promptbox}[2][]{%
  enhanced,
  breakable,
  listing only,
  colback=PromptBack,
  colframe=PromptFrame,
  boxrule=0.55pt,
  arc=2.2mm,
  left=1.8mm,
  right=1.8mm,
  top=2.8mm,
  bottom=1.4mm,
  before skip=0.8em,
  after skip=0.8em,
  listing options={
    style=PromptBoxListing
  },
  title={
    \sffamily\bfseries\color{PromptTitle}
    Prompt Name\quad
    {\normalfont\ttfamily\detokenize{#2}}
  },
  fonttitle=\small,
  colbacktitle=PromptBack,
  coltitle=PromptTitle,
  attach boxed title to top left={xshift=2mm,yshift=-2mm},
  boxed title style={
    colback=white,
    colframe=PromptFrame,
    boxrule=0.45pt,
    arc=1.8mm,
    left=1.2mm,
    right=1.2mm,
    top=0.5mm,
    bottom=0.5mm
  },
  notitle after break,
  #1
}

\newcommand{\ours}{\textit{SkillSynth}\xspace}

\title{Toward Scalable Terminal Task Synthesis via Skill Graphs}

\author{
Zhiyuan Fan \quad Tinghao Yu \quad Yuanjun Cai \quad Jiangtao Guan \quad Yun Yang \quad Dingxin Hu
\\[0.2cm]
\textbf{~Jiang Zhou \quad Xing Wu \quad Zhuo Han \quad Feng Zhang \quad Lilin Wang}
\\[0.2cm]
Hunyuan Team, Tencent\\
\Letter~ zhiyuan.fan@connect.ust.hk,~\{maxwellyu, lilinwang\}@tencent.com\\
}

\iclrfinalcopy
\begin{document}
\maketitle

\begin{abstract}
Terminal agents have demonstrated strong potential for autonomous command-line execution, yet their training remains constrained by the scarcity of high-quality and diverse execution trajectories. Existing approaches mitigate this bottleneck by synthesizing large-scale terminal task instances for trajectory sampling. However, they primarily focus on scaling the number of tasks while providing limited control over the diversity of execution trajectories that agents actually experience during training.
In this paper, we present \ours, an automated framework for terminal task synthesis built on a \textit{scenario}-mediated skill graph. \ours first constructs a large-scale skill graph, where scenarios serve as intermediate transition nodes that connect diverse command-line skills. It then samples paths from this graph as abstractions of real-world workflows, and uses a multi-agent harness to instantiate them into executable task instances. By grounding task synthesis in graph-sampled workflow paths, \ours explicitly controls the diversity of minimal execution trajectories required to solve the synthesized tasks.
Experiments on Terminal-Bench demonstrate the effectiveness of \ours. Moreover, task instances synthesized by \ours have been adopted to train Hy3 Preview, contributing to its enhanced agentic capabilities in terminal-based settings.
\end{abstract}

\section{Introduction}
Terminal agents leverage the command-line interface (CLI) as a universal action space, enabling large language models (LLMs) to execute complex, long-horizon tasks across computing systems~\citep{jimenez2024swebenchlanguagemodelsresolve,openthoughts-agent,merrill2026terminalbenchbenchmarkingagentshard}. Yet their capabilities remain fundamentally limited by the scarcity of high-quality, diverse trajectories for training. Since manually curating executable terminal tasks is expensive and difficult to scale~\citep{lin2018nl2bashcorpussemanticparser,merrill2026terminalbenchbenchmarkingagentshard}, recent work has turned to large-scale terminal task synthesis as a path toward scalable trajectory collection.

To understand what makes such trajectories useful for training, it helps to look at how terminal agents actually operate. During execution, a terminal agent does not plan over the entire environment in a single shot. Instead, it acts through step-by-step interaction with a sequence of intermediate \textit{scenarios}, applying a \textit{skill} at each scenario to make progress. An execution trajectory is therefore jointly characterized by two dimensions: the scenarios it traverses and the skills it exercises. Training terminal agents thus amounts to learning to apply appropriate skills across diverse partial scenarios, which in turn requires trajectories that are diverse along both dimensions.

Existing efforts, however, primarily scale terminal task instances, either by broadening domain coverage via LLM-generated taxonomies, which often diverge from real-world usage~\citep{pi2026dataengineeringscalingllm,zhu2026termigenhighfidelityenvironmentrobust}, or by deriving task instances from real GitHub repositories and inverting healthy environments into buggy states~\citep{wu2026largescaleterminalagentictrajectory,lin2026cligymscalableclitask,chen2026sweuniversescalerealworldverifiable}, which remain narrowly scoped to software-engineering domains such as issue resolution and feature development~\citep{yang2025swesmithscalingdatasoftware,zhang2025swebenchgoeslive,wang2025swebenchframeworkscalablegeneration}. These approaches provide limited explicit control over the scenario or skill composition underlying the resulting trajectories.

\begin{wrapfigure}{l}{0.47\textwidth}
\centering
\vspace{-0.5em}
\includegraphics[width=0.47\textwidth]{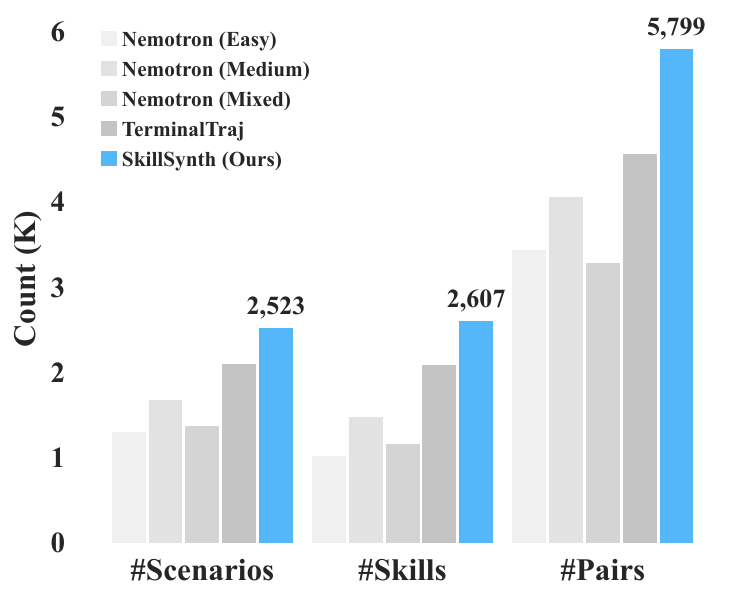}
\caption{Diversity of synthesized trajectories across datasets, measured by the number of unique scenarios, skills, and (scenario, skill) pairs after semantic canonicalization. Each value is averaged over three independent samples of 1{,}000 trajectories per dataset.}
\label{fig:diversity}
\vspace{-1.5em}
\end{wrapfigure}

As empirically shown in Figure~\ref{fig:diversity}, these trajectories exhibit redundancy in both scenario coverage and skill usage: different task instances often expose the agent to overlapping scenarios and reuse similar skills~\citep{wu2026largescaleterminalagentictrajectory}. 

\begin{figure*}[t]
    \centering
    \includegraphics[width=\textwidth]{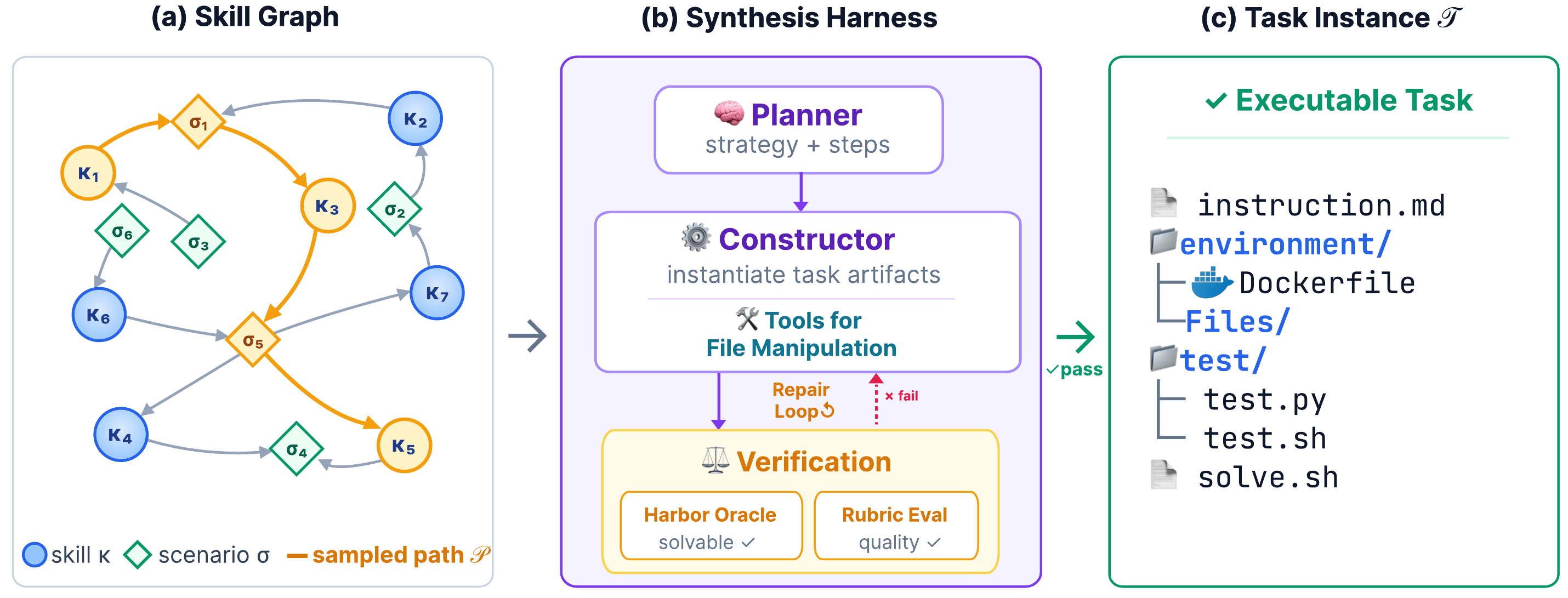}
    \caption{\textbf{Overview of \ours.} (a)~A compositional path $\mathcal{P}$ of scenarios $\sigma$ and skills $\kappa$ (highlighted in orange) is sampled from the scenario-mediated skill graph. (b)~A multi-agent harness instantiates $\mathcal{P}$ into the components of a task instance through a planner and a tool-augmented synthesis agent, followed by dual verification that checks solvability (execution-based) and specification quality (rubric-based); failed instances re-enter the synthesis loop for iterative repair. (c)~The resulting task instance $\mathcal{T}$ is fully executable and verified.}
\label{fig:main-pipeline}
\vspace{-2em}
\end{figure*}
To address this gap, we propose \ours, a scalable framework for constructing diverse terminal tasks.
\ours first collects skills from ClawHub~\citep{clawhub2026} and public GitHub repositories, capturing practical experience distilled from real terminal usage. For each skill, \ours derives a precondition scenario that describes when the skill is applicable, and a postcondition scenario that describes the state reached after its execution. It then links semantically compatible scenarios across skills to construct a scenario-mediated skill graph, in which scenarios serve as nodes and skills as directed transitions between them. A directed path sampled from the graph therefore specifies an agentic workflow, consisting of a sequence of skills ordered together with the intermediate scenarios they traverse. We then use a multi-agent harness to instantiate each sampled workflow into a concrete executable task instance whose intended solution realizes the sampled path, with oracle-based verification to ensure solvability and rubric-based evaluation to ensure task quality.

In a single fully automated run, \ours constructs 3{,}560 verified task instances from 3{,}721 sampled paths, achieving a 95.7\% oracle pass rate at only an average cost of \$27.3 per verified task instance. Compared with tasks synthesized from single-skill seeds or randomly composed multi-skill seeds, \ours tasks are more challenging and diverse: Claude Opus 4.6 requires 37 steps on average to solve them, and 121 tasks remain unsolved after three independent rollouts. We validate the effectiveness of \ours by performing supervised fine-tuning on Qwen3-8B and Qwen3-32B using sampled trajectories, and evaluating on Terminal-Bench 1.0 and 2.0. With the enhanced diversity of the sampled trajectories, the fine-tuned models achieve improved performance with higher data efficiency. Additionally, task instances synthesized by \ours have been adopted to train Hy3 Preview~\citep{hy3preview2026}, contributing to its enhanced agentic capabilities.

Currently, the constructed skill graph contains 82{,}073 scenarios after deduplication and merging, 57{,}214 filtered skills, and 185{,}529 LLM-verified bridges, from which a large space of workflow paths can be sampled for task construction. Its scenario-mediated structure makes the graph a naturally scalable infrastructure for task synthesis. As the community contributes more skills, the graph continues to expand, enabling continual synthesis of diverse terminal tasks.

Our contributions are summarized as follows:

1. We model agentic trajectories as sequences of scenarios and skills to analyze their diversity, and introduce a scenario-mediated skill graph that organizes existing skills as a foundation for sampling workflows for controllable synthesis of terminal task instances.

2. We build an end-to-end multi-agent harness that synthesizes sampled paths into executable task instances without human intervention, achieving a 95.7\% oracle pass rate and producing 3{,}560 verified task instances in a single fully automatic run.

3. We collect trajectories on the synthesized task instances and demonstrate that graph-guided synthesis yields harder tasks and more diverse trajectories than single-skill or composed multi-skill baselines, leading to consistent gains on Terminal-Bench 1.0 and 2.0 across model scales.

\section{Problem Formulation}
\label{sec:formulation}
\paragraph{Terminal agent task.}

We formulate a terminal agent task as a tuple $\tau = (\mathcal{E}, s_0, g, V)$. 
From the initial state $s_0 \in \mathcal{S}$, an agent $\pi$ interacts with the executable environment $\mathcal{E}$ to achieve a natural language described goal $g$. At step $t$ under partial observability, the agent receives an observation $o_t \in \mathcal{O}$ and samples an action $a_t \sim \pi(\cdot \mid o_{\leq t}, a_{< t}, g)$. The resulting low-level trajectory $\zeta = (o_0, a_0, \ldots, o_T)$ is labeled successful if the final state satisfies the external verifier, i.e., $V(s_T) = 1$.
\paragraph{Scenario and skill abstraction.}
However, low-level trajectories $\zeta$ consist only of fine-grained observations and actions, obscuring the high-level strategy the agent applies during execution. We thus lift $\zeta$ to a higher semantic abstraction through two objects. A \emph{scenario} $\sigma_t \in \Omega$ is a decision-relevant abstraction of the observation at a decision point of execution, and by construction serves as a sufficient statistic of the interaction history up to that point for the agent's next decision. A \emph{skill} $\kappa_t \in \mathcal{K}$ is an action subsequence applied at one scenario that produces a predictable transition to the next,
\begin{equation}
    \kappa_t : \sigma_{t-1} \to \sigma_t, \qquad \kappa_t = (a_{i_t}, a_{i_t+1}, \ldots).
    \label{eq:skill}
\end{equation}
Under this identification, $\zeta$ induces an \emph{execution trajectory}
\begin{equation}
    \xi = \big(\sigma_0, \kappa_1, \sigma_1, \kappa_2, \ldots, \kappa_L, \sigma_L\big) \in (\Omega \times \mathcal{K})^L \times \Omega,
    \label{eq:traj}
\end{equation}
which jointly characterizes the scenarios traversed and the skills exercised (cf.\ options in hierarchical RL~\citep{sutton1999between}).

\paragraph{Learning objective.}
Based on Equation~\ref{eq:traj}, an agent reduces to a policy over skills conditioned on scenarios, and training such an agent thus amounts to maximizing the expected skill-selection likelihood under the empirical distribution $\mathcal{D}$ induced by the training trajectories:
\begin{equation}
    \mathcal{J}(\pi) = \mathbb{E}_{\xi \sim \mathcal{D}} \sum_{t=1}^{L} \log \pi\!\left( \kappa_t \mid \sigma_{t-1}, g \right),
    \label{eq:objective}
\end{equation}
which is equivalent to the standard next-token-prediction loss on low-level interaction trajectories (\S~\ref{app:objective-equivalence}).
For a fixed goal $g$, decomposing $\mathcal{J}$ over the support of $\mathcal{D}$ yields
\begin{equation}
    \mathcal{J}(\pi) = \sum_{\sigma \in \Omega} p_{\mathcal{D}}(\sigma \mid g) \sum_{\kappa \in \mathcal{K}_\sigma} p_{\mathcal{D}}(\kappa \mid \sigma, g)\, \log \pi(\kappa \mid \sigma, g),
    \label{eq:decomp}
\end{equation}
where $\mathcal{K}_\sigma \subseteq \mathcal{K}$ denotes skills admissible at $\sigma$. Equation~\ref{eq:decomp} makes explicit that the learnable region of $\pi$ is confined to the support of $\mathcal{D}$ along both factors: scenarios with $p_\mathcal{D}(\sigma \mid g) = 0$ are unobservable, and skills with $p_\mathcal{D}(\kappa \mid \sigma, g) = 0$ are unexercised. Maximizing learned capacity therefore requires training data whose induced $\mathcal{D}$ densely covers the conditional product space $\{(\sigma, \kappa): \sigma \in \Omega,\, \kappa \in \mathcal{K}_\sigma\}$, which motivates our approach to scale synthetic task instances by maximizing the diversity of execution trajectories experienced by the agent.

\section{Approach}
\label{sec:method}

\subsection{Overview}
\ours synthesizes diverse terminal task instances through three stages. We first construct a scenario-mediated skill graph, in which nodes are scenarios and directed edges represent skills that point from precondition to postcondition scenarios (\S~\ref{subsec:skill-graph-construction}). From this graph we sample directed paths, each specifying a compositional sequence of skills to be exercised sequentially together with the scenarios they traverse (\S~\ref{subsec:graph-guided-task-synthesis}). A multi-agent harness then instantiates each sampled path into a concrete, executable task instance with oracle verification and rubric-based evaluation to ensure solvability and task quality (\S~\ref{subsec:multi-agent-harness}).

\begin{figure}
    \centering
    \includegraphics[width=1\linewidth]{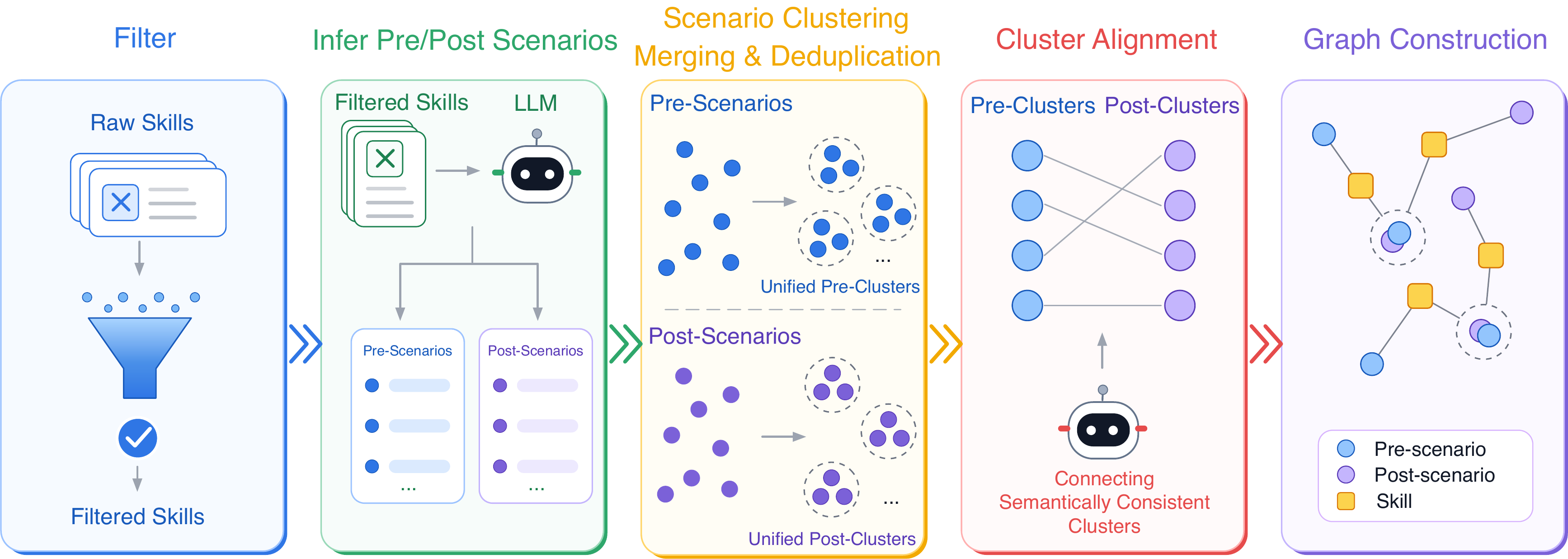}
    \caption{Overview of the skill graph construction pipeline.}
    \label{fig:skill-graph-pipeline}
    \vspace{-1em}
\end{figure}

\subsection{Skill Graph Construction}
\label{subsec:skill-graph-construction}
Recall from \S~\ref{sec:formulation} that each skill $\kappa \in \mathcal{K}$ is a directed transition $\kappa: \sigma \to \sigma'$ between scenarios. We lift this pairwise structure into a directed multigraph $\mathcal{G} = (\Omega, \mathcal{K})$, where nodes are scenarios and edges are skills. A directed path in $\mathcal{G}$ corresponds to a sequential workflow in which each skill's postcondition serves as the precondition of the next. The full construction pipeline is illustrated in Figure~\ref{fig:skill-graph-pipeline}.

\textbf{Skill filtering.} We instantiate $\mathcal{G}$ from human-written skills in ClawHub~\citep{clawhub2026} and public GitHub repositories. The skill pool $\mathcal{K}$ retains only skills that are 1)~executable on a Linux terminal, 2)~defined by a structured workflow rather than prompt engineering alone, 3)~free of adversarial or jailbreak content (e.g., downloading files from unknown IPs, exfiltrating environment keys), and 4)~producing deterministic, objectively verifiable outputs.

\textbf{Scenario inference.} For each retained skill $\kappa \in \mathcal{K}$, we prompt an LLM with its full specification (Markdown description, code, and usage examples) to infer plausible precondition scenarios $\Omega_\kappa^{\mathrm{pre}}$ and postcondition scenarios $\Omega_\kappa^{\mathrm{post}}$, yielding atomic transitions $\{\kappa : \sigma \to \sigma' \mid \sigma \in \Omega_\kappa^{\mathrm{pre}},\ \sigma' \in \Omega_\kappa^{\mathrm{post}}\}$.

\textbf{Scenario deduplication.} All inferred scenarios are embedded and undergo clustering-based semantic deduplication, merging scenarios that describe the same state but differ in lexical form. We evaluate nine common clustering algorithms and find that a hierarchical agglomerative clustering method with Louvain-based coarse bucketing performs best empirically.

\textbf{Cross-skill alignment.} To connect atomic transitions into a unified graph, postconditions of one skill are aligned with preconditions of another. For each postcondition scenario, the top-1{,}000 most similar precondition scenarios are retrieved by embedding similarity, and an LLM judges semantic compatibility. We repeat this process in reverse (precondition $\to$ top-1{,}000 postconditions) using separately designed prompts to ensure bidirectional alignment quality.

\textbf{Scenario merging and filtering.} Aligned pre- and postcondition pairs are merged into unified scenario nodes using an LLM. Finally, we perform an LLM-based filtering pass over all resulting (scenario, skill, scenario) triples, retaining only those that form valid transitions. Through manual review of sampled graph cases, we find that each stage is necessary to ensure the overall quality of the constructed graph. Full alignment criteria and prompt details are provided in \S~\ref{app:graph_discussion}.

\subsection{Graph-Guided Path Sampling}
\label{subsec:graph-guided-task-synthesis}

Given the skill graph $\mathcal{G} = (\Omega, \mathcal{K})$, we sample directed paths that serve as compositional inputs to the multi-agent synthesis harness (\S~\ref{subsec:multi-agent-harness}). A sampled path
\[
\mathcal{P} = (\sigma_0, \kappa_1, \sigma_1, \kappa_2, \ldots, \kappa_L, \sigma_L)
\]
interleaves $L$ skill transitions with $L+1$ scenarios, where each $\kappa_l \in \mathcal{K}$ is a directed edge from $\sigma_{l-1}$ to $\sigma_l$ in $\mathcal{G}$. Paths whose length $L$ falls within $[L_{\min}, L_{\max}]$ are retained, covering single-skill tasks ($L \in \{1, 2, 3\}$) through compositional multi-step ones ($L \geq 4$); we set $L_{\min} = 1$ and $L_{\max} = 7$ in our experiments.

A uniform random walk on $\mathcal{G}$ concentrates on high-degree scenarios and frequently traversed skills, producing redundant paths that revisit the same sub-structures. We instead sample paths with inverse-frequency weighting: for each scenario $\sigma \in \Omega$ we maintain a visit count $\nu(\sigma)$, and for each skill $\kappa \in \mathcal{K}$ a usage count $\mu(\kappa)$. The source scenario is drawn with probability $p(\sigma) \propto (\nu(\sigma) + 1)^{-1}$, and at each step of the walk the next edge is drawn from the outgoing skills of the current node with probability $p(\kappa) \propto (\mu(\kappa) + 1)^{-1}$. To enforce monotone progression, visited scenarios and skills are excluded from subsequent steps within the same path. The walk continues until it reaches $L_{\max}$ or encounters a dead-end with no valid continuation. If the resulting path length falls within $[L_{\min}, L_{\max}]$ and its skill set has not been seen before, the path is accepted and both counters are incremented, progressively steering the empirical distribution toward uniform coverage over $\Omega \times \mathcal{K}$, which is the coverage criterion derived in Equation~\ref{eq:decomp}. The full procedure is given in Algorithm~\ref{alg:graph-sampling}.

\begin{algorithm}[t]
\caption{Inverse-Frequency Path Sampling with Monotone Progression}
\label{alg:graph-sampling}
\begin{algorithmic}[1]
\REQUIRE Skill graph $\mathcal{G} = (\Omega, \mathcal{K})$, length range $[L_{\min}, L_{\max}]$, sampling budget $N$
\ENSURE Set of unique sampled paths $\Pi$
\STATE Initialize $\nu(\sigma) \leftarrow 0$ for all $\sigma \in \Omega$,\; $\mu(\kappa) \leftarrow 0$ for all $\kappa \in \mathcal{K}$,\; $\Pi \leftarrow \emptyset$,\; $\mathcal{S} \leftarrow \emptyset$
\FOR{$b = 1, \ldots, N$}
    \STATE Sample $\sigma_0 \in \Omega$ with probability $\propto (\nu(\sigma) + 1)^{-1}$
    \STATE $\mathcal{P} \leftarrow (\sigma_0)$,\; $\mathcal{V}_\kappa \leftarrow \emptyset$,\; $\mathcal{V}_\sigma \leftarrow \{\sigma_0\}$,\; $l \leftarrow 0$
    \WHILE{$l < L_{\max}$}
        \STATE $\mathcal{N}(\sigma_l) \leftarrow \bigl\{\kappa \in \mathcal{K} \;:\; \kappa\colon \sigma_l \to \sigma' \text{ for some } \sigma' \in \Omega \setminus \mathcal{V}_\sigma,\; \kappa \notin \mathcal{V}_\kappa\bigr\}$
        \IF{$\mathcal{N}(\sigma_l) = \emptyset$}
            \STATE \textbf{break} \COMMENT{dead-end; no monotone continuation}
        \ENDIF
        \STATE Sample $\kappa_{l+1} \in \mathcal{N}(\sigma_l)$ with probability $\propto (\mu(\kappa) + 1)^{-1}$
        \STATE Sample $\sigma_{l+1} \in \kappa_{l+1}.\mathrm{post} \setminus \mathcal{V}_\sigma$ with probability $\propto (\nu(\sigma) + 1)^{-1}$
        \STATE Append $(\kappa_{l+1}, \sigma_{l+1})$ to $\mathcal{P}$
        \STATE $\mathcal{V}_\kappa \leftarrow \mathcal{V}_\kappa \cup \{\kappa_{l+1}\}$,\; $\mathcal{V}_\sigma \leftarrow \mathcal{V}_\sigma \cup \{\sigma_{l+1}\}$
        \STATE $l \leftarrow l + 1$
    \ENDWHILE
    \IF{$L_{\min} \leq l \leq L_{\max}$ \AND $\mathrm{skills}(\mathcal{P}) \notin \mathcal{S}$}
        \STATE $\Pi \leftarrow \Pi \cup \{\mathcal{P}\}$;\; $\mathcal{S} \leftarrow \mathcal{S} \cup \{\mathrm{skills}(\mathcal{P})\}$
        \STATE $\nu(\sigma) \leftarrow \nu(\sigma) + 1$ for each $\sigma$ in $\mathcal{P}$
        \STATE $\mu(\kappa) \leftarrow \mu(\kappa) + 1$ for each $\kappa$ in $\mathcal{P}$
    \ENDIF
\ENDFOR
\RETURN $\Pi$
\end{algorithmic}
\end{algorithm}

\subsection{Multi-Agent Harness}
\label{subsec:multi-agent-harness}
Given a sampled path $\mathcal{P}$ from \S~\ref{subsec:graph-guided-task-synthesis}, a multi-agent harness is designed to produce an executable task instance consisting of five components: a natural-language instruction, an initial filesystem snapshot, a containerized environment, verification scripts, and an oracle solution.

\begin{figure}[t]
    \centering
    \includegraphics[width=1\linewidth]{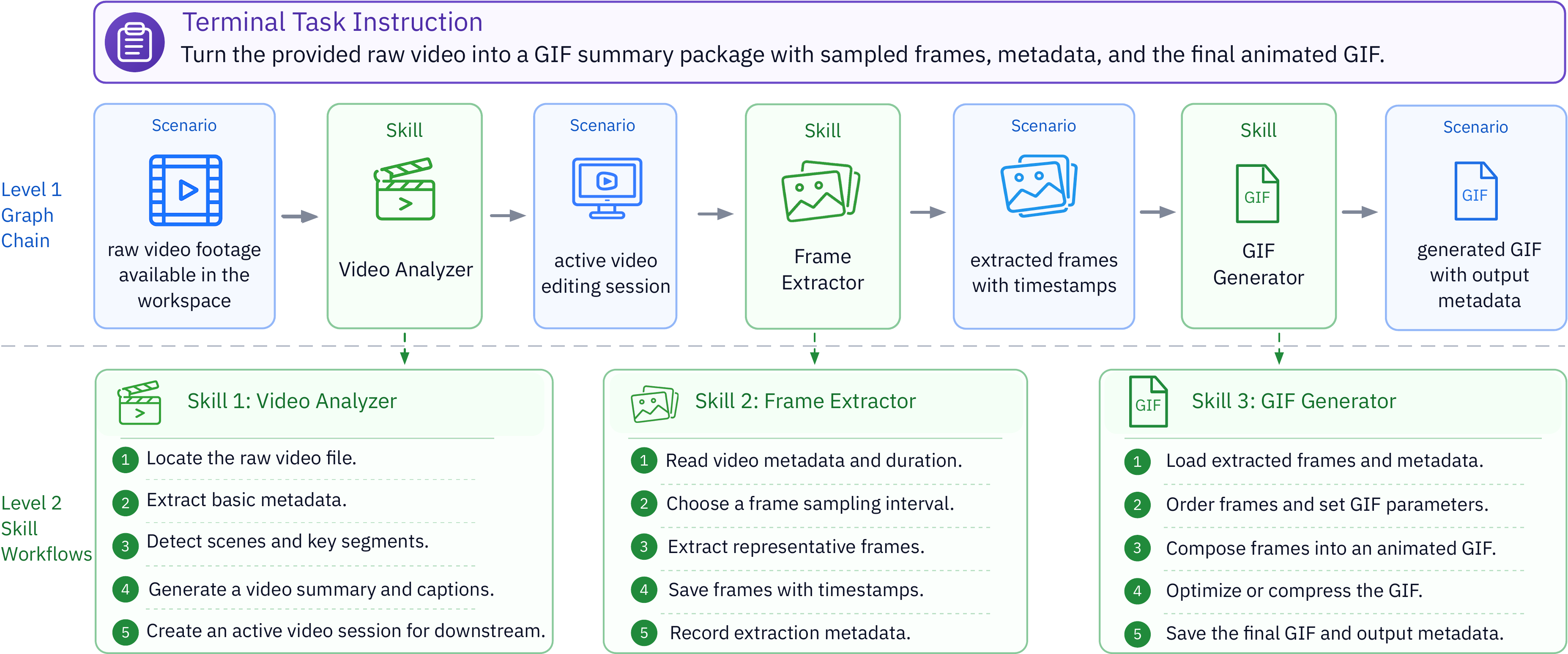}
    \caption{
    A video-domain example of a path sampled from the skill graph and the corresponding synthesized task instruction.
    To accomplish the task, the agent needs to apply multiple skills, each of which expands into a multi-step internal workflow.
    }
    \label{fig:example}
\end{figure}

Directly prompting an LLM to generate all components in a single pass leads to two issues: 1)~long-context generation produces outputs of inconsistent quality, and 2)~the model focuses on implementation details rather than designing a coherent task instance, resulting in tasks that lack sufficient complexity. We thus decouple planning from implementation: a planner first transforms $\mathcal{P}$ into a structured plan of sub-objectives and expected outputs, then a constructor generates the whole task instance conditioned on this plan. Each task instance is checked along two complementary axes: \emph{execution-based verification} runs the verification scripts against the oracle solution inside the environment container to ensure solvability; and \emph{rubric-based verification} uses LLM-as-a-Judge to assess (i)~alignment between instructions and tests, ensuring that tests are neither too lenient (missing required functionality) nor too strict (imposing unstated constraints), and (ii)~instruction self-containedness, ensuring that no hints about the oracle solution progress leak into the instruction. If either check fails, diagnostic feedback is returned to the constructor for repair via multi-turn tool use. We permit up to $R = 3$ repair cycles, each with at most $N_{\mathrm{tool}} = 20$ tool calls, after which the instance is accepted or discarded. An example of a sampled path and the corresponding synthesized task instruction is shown in Figure~\ref{fig:example}.

\section{Experiments}

\subsection{Experimental Setup}
We sample 3{,}721 paths from the constructed skill graph to synthesize task instances and validate the effectiveness of \ours.

\paragraph{Base Model.}
We select the Qwen3 dense series~\citep{yang2025qwen3technicalreport} (8B, 14B, and 32B) as our base models to study the effect of model scale, with Qwen3-32B serving as our primary model for ablation studies. All models are trained with full-parameter supervised fine-tuning using a learning rate of 2e-5 for 5 epochs. Full training details are provided in \S~\ref{app:sft}.

\paragraph{Evaluation.}
We use Terminal-Bench 1.0 and 2.0~\citep{merrill2026terminalbenchbenchmarkingagentshard} as our evaluation benchmarks, comprising 80 and 89 community-curated tasks, respectively, with the latter serving as a harder successor to the former. All reported results are the mean accuracy over three independent runs, along with 95\% confidence intervals.
We adopt Terminus 2 as our agent scaffold, which interacts with the environment solely through a single headless terminal without additional harness constraints. For evaluation infrastructure, we employ Harbor~\citep{Harbor_Framework_Team_Harbor_A_framework_2026}, the official orchestration framework of Terminal-Bench 2.0, to parallelize trajectory sampling across 128 concurrent Docker~\citep{merkel2014docker} environments for enhanced sampling efficiency.

\subsection{Multi-agent Harness Quality}
\begin{table}[t]
\centering
\begin{minipage}{0.48\textwidth}
\centering
\small
\begin{tabular}{lcc}
\toprule
\multicolumn{3}{c}{\textbf{Task Verification Results}} \\
\midrule
\textbf{Category} & \textbf{Count} & \textbf{Ratio (\%)} \\
\midrule
All passed                   & 3,423 & 92.0 \\
Oracle passed only                  & 137   & 3.7 \\
Failed                       & 161   & 4.3 \\
\midrule
Total                        & 3,721 & 100.0 \\
\midrule
\multicolumn{3}{c}{\textbf{Repair Statistics}} \\
\midrule
Avg. repair cycles & \multicolumn{2}{c}{2.31} \\
Avg. tool calls    & \multicolumn{2}{c}{11} \\
Recovered tasks    & \multicolumn{2}{c}{721} \\
\bottomrule
\end{tabular}
\caption{Summary of multi-agent harness outcomes and repair statistics.}
\label{tab:harness-summary}
\end{minipage}
\hfill
\begin{minipage}{0.48\textwidth}
\centering
\small
\begin{tabular}{lcc}
\toprule
\textbf{Successes (out of 3)} & \textbf{Percentage} & \textbf{\# Tasks} \\
\midrule
0/3 & 38\% & 1{,}352 \\
1/3 & 18\% & 637 \\
2/3 & 19\% & 679 \\
3/3 & 25\% & 892 \\
\midrule
Total & 100\% & 3{,}560 \\
\bottomrule
\end{tabular}
\caption{Difficulty distribution of the 3{,}560 usable task instances. Lower success counts indicate harder tasks.}
\label{tab:difficulty-dist}
\end{minipage}
\end{table}

\textbf{The multi-agent harness achieves high yield.} As shown in Table~\ref{tab:harness-summary}, 95.7\% of synthesized instances pass the oracle check, with 92.0\% passing both quality checks. The multi-turn verify-then-repair loop recovers 721 task instances that failed in the first round, which shows that interactive repair is necessary to achieve high yield at scale. In total, a single fully automatic run produces 3{,}560 usable task instances. Scaling further simply requires sampling more paths from the skill graph and re-running the multi-agent harness.

\textbf{Misalignment between instructions and tests dominates rubric failures.} Of failed rubric checks, 77\% stem from test scripts that either over-specify or under-specify relative to the instruction, potentially producing inaccurate evaluation of agent trajectories and erroneous reward signals for reinforcement learning. We retain these task instances for supervised fine-tuning to preserve trajectory diversity, but discard them for reinforcement learning to avoid erroneous reward signals.

\textbf{Poor first-round generations are hard to recover.} For oracle failures, the dominant cause is corrupted filesystem snapshots: due to the randomness of LLM generation, the first synthesis round may produce buggy stages that remain unrecoverable even after multiple repair cycles. Although the multi-agent architecture substantially outperforms a single-stage pipeline, these cases suggest a floor on repair effectiveness. Re-running failed paths with a higher sampling temperature is a simple yet effective strategy to help recover these paths and preserve the diversity of the synthesized task set.

\textbf{Task difficulty distribution.} We categorize the 3{,}560 usable task instances by difficulty using Hy3 Preview~\citep{hy3preview2026}. Each task is attempted three times, and we assign difficulty levels based on the number of successful attempts. Table~\ref{tab:difficulty-dist} reports the distribution. 37\% of tasks fall in the learnable range (1/3 or 2/3 success rate), while 38\% (0/3) represent potentially challenging problems for future exploration.

\subsection{Main Results}
\begin{table}[t]
\centering
\small
\begin{tabular}{lcc c lcc}
\toprule
\multicolumn{3}{c}{\textbf{Proprietary}} & & \multicolumn{3}{c}{\textbf{Open-Source}} \\
\cmidrule{1-3} \cmidrule{5-7}
\textbf{Model} & \textbf{TB 1.0} & \textbf{TB 2.0} & & \textbf{Model} & \textbf{TB 1.0} & \textbf{TB 2.0} \\
\midrule
GPT-5.3-Codex    & --             & 64.7 $\pm$ 2.7 & & GLM 5             & --             & 52.4 $\pm$ 2.6 \\
Claude Opus 4.6   & --             & 62.9 $\pm$ 2.7 & & Kimi K2.5         & --             & 43.2 $\pm$ 2.9 \\
Claude Opus 4.5   & --             & 57.8 $\pm$ 2.5 & & MiniMax m2.5      & --             & 42.2 $\pm$ 2.6 \\
Gemini 3 Pro      & --             & 56.9 $\pm$ 2.5 & & Grok 4            & 39.0 $\pm$ 3.2 & --             \\
GPT-5.2           & --             & 54.0 $\pm$ 2.9 & & DeepSeek-V3.2     & --             & 39.6 $\pm$ 2.8 \\
Claude Sonnet 4.5 & 51.0 $\pm$ 1.6 & 42.8 $\pm$ 2.8 & & Grok 4 Fast       & 31.3 $\pm$ 2.8 & --             \\
GPT-5.1           & --             & 47.6 $\pm$ 2.8 & & Qwen 3 Coder 480B & --             & 23.9 $\pm$ 2.8 \\
Claude Opus 4.1   & 43.8 $\pm$ 2.8 & 38.0 $\pm$ 2.6 & &                   &                &                \\
Claude Haiku 4.5  & 41.8 $\pm$ 2.6 & 28.3 $\pm$ 2.9 & &                   &                &                \\
GPT-5             & 41.3 $\pm$ 2.2 & 35.2 $\pm$ 3.1 & & \multicolumn{3}{c}{\textbf{Ours}} \\
\cmidrule{5-7}
Claude Opus 4     & 39.0 $\pm$ 0.8 & --             & & Qwen3-8B + SS     & 17.1 $\pm$ 1.8 & 13.5 $\pm$ 2.8 \\
Claude Sonnet 4   & 36.4 $\pm$ 1.2 & --             & & Qwen3-14B + SS    & 22.9 $\pm$ 1.8 & 19.9 $\pm$ 1.6 \\
GPT-5-Mini        & 30.8 $\pm$ 3.9 & 24.0 $\pm$ 2.5 & & Qwen3-32B + SS    & 33.8 $\pm$ 3.1 & 29.6 $\pm$ 1.6 \\
\bottomrule
\end{tabular}
\caption{Experimental results on Terminal-Bench 1.0 and 2.0 with the Terminus 2 scaffold. +SS denotes models trained with \ours generated trajectories.}
\label{tab:terminal-bench-main}
\end{table}
For trajectory collection, we use MiniMax M2.7 as the teacher model due to its cost efficiency, sampling three trajectories per task instance for a total of 10{,}680 trajectories. We retain both successful and failed trajectories for training~\citep{pi2026dataengineeringscalingllm} to preserve diversity and coverage, as failed trajectories contain useful reasoning patterns for agentic problem-solving, such as error diagnosis and recovery strategies across diverse scenarios.

Table~\ref{tab:terminal-bench-main} reports experimental results after training Qwen3-8B, Qwen3-14B, and Qwen3-32B on collected trajectories. All three models improve over their respective baselines, with gains scaling with model size. Qwen3-32B + SS outperforms the larger Qwen 3 Coder 480B on TB 2.0, suggesting that targeted data construction and domain-specific training can effectively improve the terminal agentic capabilities of smaller models.

\subsection{Ablation Study}

To isolate the contributions of skill graph construction and path sampling, we compare \ours against two baselines that use the same multi-agent harness for task synthesis. We first construct a skill pool from all skills appearing in the sampled paths, then create: 1) Single-skill: 3{,}721 skills randomly drawn from the pool, each used as a standalone seed for task synthesis. 2) Multi-skill: 3{,}721 randomly composed combinations of 2--7 skills from the pool, without graph-guided ordering.

\begin{table}[h]
\centering
\small
\begin{tabular}{lcccccc}
\toprule
& \multicolumn{4}{c}{Task Distribution} & \multicolumn{2}{c}{Accuracy} \\
\cmidrule(lr){2-5} \cmidrule(lr){6-7}
Strategy & 0/3 & 1/3 & 2/3 & 3/3 & TB 1.0 & TB 2.0 \\
\midrule
Single-skill    & 16\% & 23\% & 34\% & 27\% & 25.4 $\pm$ 1.8 & 21.3 $\pm$ 2.8 \\
Multi-skill     & 27\% & 24\% & 21\% & 28\% & 30.8 $\pm$ 1.8 & 25.8 $\pm$ 2.8 \\
\ours~(Ours)    & 38\% & 18\% & 19\% & 25\% & 33.8 $\pm$ 3.1 & 29.6 $\pm$ 1.6 \\
\bottomrule
\end{tabular}
\caption{Ablation study on synthesis seed selection strategy.}
\label{tab:ablation}
\end{table}
Table~\ref{tab:ablation} reports both task difficulty and downstream model performance. \ours tasks are harder than both baselines. On the training side, models trained on \ours trajectories outperform those trained on single-skill trajectories by 8.4 points on TB~1.0 and 8.3 points on TB~2.0, and outperform multi-skill trajectories by 3.0 points on TB~1.0 and 3.8 points on TB~2.0.

\textbf{Random composition lacks workflow coherence.} We find multi-skill baselines produce lower-quality tasks because randomly composed skills lack sequential dependencies and cannot form coherent workflows. The multi-agent harness tends to generate simplified task instances that contain multiple fine-grained requirements but require few execution steps.

\subsection{Diversity Analysis}
We analyze diversity at two levels: 1) the structural diversity of the skill graph itself and sampled paths, and 2) the diversity of the collected trajectories.

\textbf{Skill graph and sampled paths.} The constructed skill graph spans a wide range of domains, including long-tail areas such as Audio \& Speech and 3D \& Simulation (see \S~\ref{app:graph_analysis} for the full domain distribution). The graph is highly connected at the macro level: 85.6\% of scenarios belong to the largest weakly connected component, while 6251 smaller components capture specialized, self-contained workflows. From the whole graph, we enumerate 16{,}632{,}220 paths requiring seven or more skills, confirming a vast combinatorial space for task synthesis. Algorithm~\ref{alg:graph-sampling} ensures that the sampled paths are diverse at the skill and scenario levels.

\textbf{Trajectories.} To quantify execution diversity, we randomly sample 1{,}000 trajectories separately and use deepseek-reasoner~(v3.2) to extract the scenarios and skills encountered by the agent during execution. Using the same prompt and model across all strategies ensures statistically comparable extraction granularity.
We then embed all extracted scenarios and skills using Harrier-OSS-v1-27B and perform clustering-based deduplication to obtain unique scenarios, skills, and scenario-skill pairs. Trajectories sampled from \ours tasks exhibit 31\% higher unique scenario-skill coverage than single-skill baselines and 19\% higher than randomly composed multi-skill baselines on average. The improved diversity translates directly into higher data efficiency and performance, as shown in Table~\ref{tab:ablation}. Prompts are provided in \S~\ref{app:div_prompt}.

\subsection{Error Analysis}

\begin{table}[h]
\centering
\small
\begin{tabular}{lcp{8cm}}
\toprule
\textbf{Error Category} & \textbf{Ratio} & \textbf{Typical Example} \\
\midrule
Partial Implementation      & 42.2\% & Agent runs \texttt{pytest} on self-generated tests, gets all tests passing, and submits but silently skips one or more clauses from the instruction. \\
\midrule
Inline Self-Test Over-trust & 29.5\% & Agent never runs \texttt{pytest}; instead verifies with hand-picked \texttt{python3 -c} assertions on happy paths, missing edge cases. \\
\midrule
Premature Termination       & 12.9\% & Agent finishes writing code and exits without running any form of verification. \\
\midrule
API/Flag Hallucination      & 7.0\%  & Agent repeatedly invokes non-existent modules or flags, triggering $\geq$3 errors such as \texttt{No module named} or \texttt{unexpected kwarg}. \\
\midrule
Debug Fixation Loop         & 5.1\%  & Agent repeats the same command $\geq$10 times with minor edits, never switching strategy, and exhausts the episode budget. \\
\midrule
Error Rationalization       & 3.3\%  & Agent observes failing tests but rationalizes them as ``pre-existing'' or ``flaky'' and completes anyway. \\
\bottomrule
\end{tabular}
\caption{Error analysis of failed trajectories.}
\label{tab:error-analysis}
\end{table}
To understand failure modes, three of the authors independently analyzed 20 failed trajectories each. Based on the results, we design an error analysis skill for an LLM agent, which then analyzes all failed trajectories accordingly. As shown in Table~\ref{tab:error-analysis}, the dominant failure modes are partial implementation (42.2\%) and over-trust in inline self-test (29.5\%). In both cases, the agent substitutes self-narrated verification for specification-grounded testing and confidently marks the task complete. These findings suggest that terminal agents require stronger alignment with task instructions and more flexible exploration strategies during execution.

\subsection{Discussion and Future Work}

\ours synthesizes a large number of task instances that reflect real-world workflows, covering diverse scenarios and skills. Future work includes 1) scaling \ours to larger data regimes while continuously refining the multi-agent harness to adapt to more cost-effective models, 2) sampling subgraphs rather than chains from the skill graph, which requires parallel execution of multiple skills and increases task complexity.

\section{Related Work}
\paragraph{Terminal Agents.} Recent work has increasingly explored LLMs as terminal agents that interact with real-world computing systems through command-line interfaces~\citep{yang2023intercodestandardizingbenchmarkinginteractive,jimenez2024swebenchlanguagemodelsresolve}. To evaluate such capabilities, Terminal-Bench~\citep{merrill2026terminalbenchbenchmarkingagentshard} provides a set of hand-crafted tasks spanning diverse domains, requiring agents to complete end-to-end workflows within containerized Docker environments. Meanwhile, a series of agent scaffolds~\citep{anthropic2025claudecode,openai2025codex,google2025geminicli,wang2025openhandsopenplatformai,jetbrains2025junie} have been developed to enhance LLMs' planning, execution, and tool calling capabilities in terminal settings. Despite strong proprietary results, open-source models remain substantially behind, motivating scalable synthesis of diverse terminal execution trajectories for effective training.

\paragraph{Synthetic Data for Terminal Agents.} To reduce the cost of curating terminal training data, prior efforts scale synthetic task instances along different dimensions. Some approaches expand domain coverage through LLM-generated taxonomies, such as Endless Terminals~\citep{gandhi2026endlessterminalsscalingrl}, TermiGen~\citep{zhu2026termigenhighfidelityenvironmentrobust}, and Nemotron-Terminal~\citep{pi2026dataengineeringscalingllm}, enabling task instance generation across diverse domains. Others scale the number of task instances, for example by collecting real-world Docker environments from GitHub~\citep{wu2026largescaleterminalagentictrajectory}, or by inverting healthy environments into buggy states to derive new task instances~\citep{lin2026cligymscalableclitask}. While these approaches increase the number of task instances, they provide limited control over the diversity of execution trajectories experienced by agents during task solving. We address this limitation by synthesizing task instances from a skill graph, enabling explicit control over the trajectories encountered during training.

\paragraph{Skill Topology.}
A separate line of work organizes reusable agent skills into structured topologies, following the reusable skill library introduced by Voyager~\citep{wang2023voyageropenendedembodiedagent} and drawing on large skill repositories such as ClawHub~\citep{clawhub2026}. AgentSkillOS~\citep{li2026organizingorchestratingbenchmarkingagent} organizes skills into a hierarchical capability tree and composes them through DAG-based orchestration graphs, while SkillNet~\citep{liang2026skillnetcreateevaluateconnect} models skills as nodes in a relational graph with explicit inter-skill connections. However, these approaches primarily focus on skill organization and retrieval, rather than modeling how skills are composed during execution. In contrast, we organize skills through scenarios, where skill transitions are induced by shared executable states and temporal ordering in trajectories, enabling structured synthesis of terminal tasks with controllable scenario and skill diversity.

\section{Conclusion}
In this paper, we model agentic trajectories as sequences of scenarios and skills to analyze their diversity, and propose \ours, a framework that synthesizes diverse terminal task instances by sampling compositional paths from a constructed skill graph. \ours achieves a high synthesis pass rate at relatively low cost, enabling scalable production of verified task instances. Experimental results show that training effectiveness depends on trajectory diversity, not merely on task volume.
As communities like ClawHub continue to grow, \ours provides a scalable way to transform emerging workflows into diverse task instances, enabling models to continuously learn real-world experience.

\section*{Ethics Statement}
The skills used to construct our data are sourced from publicly available GitHub repositories and ClawHub, and are used in accordance with their respective open-source licenses.

\bibliography{iclr2026_conference}
\bibliographystyle{iclr2026_conference}

\appendix
\newpage
\section{Proof of Equivalence}
\label{app:objective-equivalence}

We show that the skill-level objective in Equation~\ref{eq:objective} reduces to the standard next-action objective under three assumptions: (A1)~the mapping $\zeta \mapsto \xi$ is deterministic, (A2)~each scenario $\sigma_t$ is a sufficient statistic for the agent's next decision, and (A3)~each skill $\kappa_t = (a_{i_t}, \ldots, a_{j_t})$ is executed autoregressively.

Under (A2) and (A3), the chain rule of conditional probability gives
\begin{equation}
    \log \pi(\kappa_t \mid \sigma_{t-1}, g) = \sum_{k=i_t}^{j_t} \log \pi(a_k \mid o_{\leq k}, a_{<k}, g).
\end{equation}
Summing over $t = 1, \ldots, L$ and using the contiguity $i_{t+1} = j_t + 1$, the double sum collapses into a single sum over all actions. Taking expectations and applying (A1) yields
\begin{equation}
    \mathcal{J}(\pi) = \mathbb{E}_{\xi \sim \mathcal{D}} \sum_{t=1}^{L} \log \pi(\kappa_t \mid \sigma_{t-1}, g) = \mathbb{E}_{\zeta \sim \mathcal{D}} \sum_{t=0}^{T-1} \log \pi(a_t \mid o_{\leq t}, a_{<t}, g).
\end{equation}
The equivalence shows that the scenario--skill abstraction imposes no additional training machinery, and properties of $\mathcal{D}$ established at the scenario--skill level (e.g., the coverage criterion in Equation~\ref{eq:decomp}) transfer directly to the token-level loss.

\section{Skill Graph Construction Discussion}
\label{app:graph_discussion}
All LLM calls use DeepSeek Reasoner~(v3.2) to ensure extraction quality and reliability. We also explored several alternative graph construction strategies, including embedding-based scenario alignment and single-pass subgraph generation; we found that the LLM-based pairwise alignment approach yields higher-quality graphs.

We cluster scenarios using a two-stage scalable hierarchical procedure. First, we construct a sparse semantic similarity graph over normalized scenario embeddings and apply Louvain community detection~\citep{Blondel_2008} to obtain coarse buckets. Then, within each bucket, we run complete-linkage agglomerative clustering~\citep{johnson1967hierarchical} with cosine distance to avoid the quadratic memory cost of global hierarchical clustering while preserving the desirable property of complete linkage: every pair within a cluster must be mutually close, preventing chain-drift artifacts where transitively linked but semantically distant scenarios get merged (e.g., $A$ is close to $B$ and $B$ is close to $C$, but $A$ and $C$ are semantically different.).

For the clustering hyperparameters, we sweep the agglomerative distance threshold on held-out scenario samples and manually inspect the resulting clusters. We choose the final threshold by balancing merge quality and over-fragmentation: the selected value should merge clear paraphrases and near-equivalent states while keeping semantically distinct states, especially negations and pre/post condition changes, in separate clusters.

\section{Supervised Fine-Tuning}
\label{app:sft}
We perform supervised fine-tuning with AdamW, using $\beta_1=0.9$ and $\beta_2=0.95$, together with a cosine learning rate schedule and a warmup ratio of $10\%$. The peak learning rate is set to $2\times10^{-5}$ and the weight decay is $1\times10^{-4}$ unless otherwise noted. We train all models for $5$ epochs in bfloat16 precision. To stabilize optimization on long terminal trajectories, we apply gradient clipping with a maximum norm of $1.0$. Training uses a micro-batch size of $1$ per GPU and gradient accumulation to achieve the desired global batch size.

\section{Skill Graph Analysis}
\label{app:graph_analysis}

\begin{table}[h]
\centering
\begin{minipage}[t]{0.52\linewidth}
\centering
\begin{tabular}{lr}
\toprule
\textbf{Metric} & \textbf{Value} \\
\midrule
Scenario nodes $|\mathcal{S}|$ & 82{,}073 \\
Skill-labeled transitions $|\mathcal{E}|$ & 57{,}214 \\
Source-only scenarios & 18{,}749 (22.8\%) \\
Sink-only scenarios & 12{,}299 (15.0\%) \\
Bridge scenarios & 46{,}699 (56.9\%) \\
\midrule
Mean / Median / Max degree & 4.32 / 2 / 752 \\
Connected components & 6{,}251 \\
Giant component & 118{,}806 (85.6\%) \\
\bottomrule
\end{tabular}
\end{minipage}%
\hfill
\begin{minipage}[t]{0.38\linewidth}
\centering
\begin{tabular}{rr}
\toprule
\textbf{Component size} & \textbf{Count} \\
\midrule
2 & 482 \\
3--5 & 5{,}561 \\
6--10 & 183 \\
11--50 & 24 \\
$>$50 & 1 \\
\midrule
\textbf{Total} & \textbf{6{,}251} \\
\bottomrule
\end{tabular}
\end{minipage}
\caption{Left: Structural statistics of the skill graph $\mathcal{G}$. Right: Connected component size distribution.}
\label{tab:graph-stats}
\end{table}

Table~\ref{tab:graph-stats} summarizes the structural properties of the skill graph $\mathcal{G}$ constructed in \S~\ref{subsec:skill-graph-construction}.

\begin{figure}[h]
    \centering
    \includegraphics[width=0.85\linewidth]{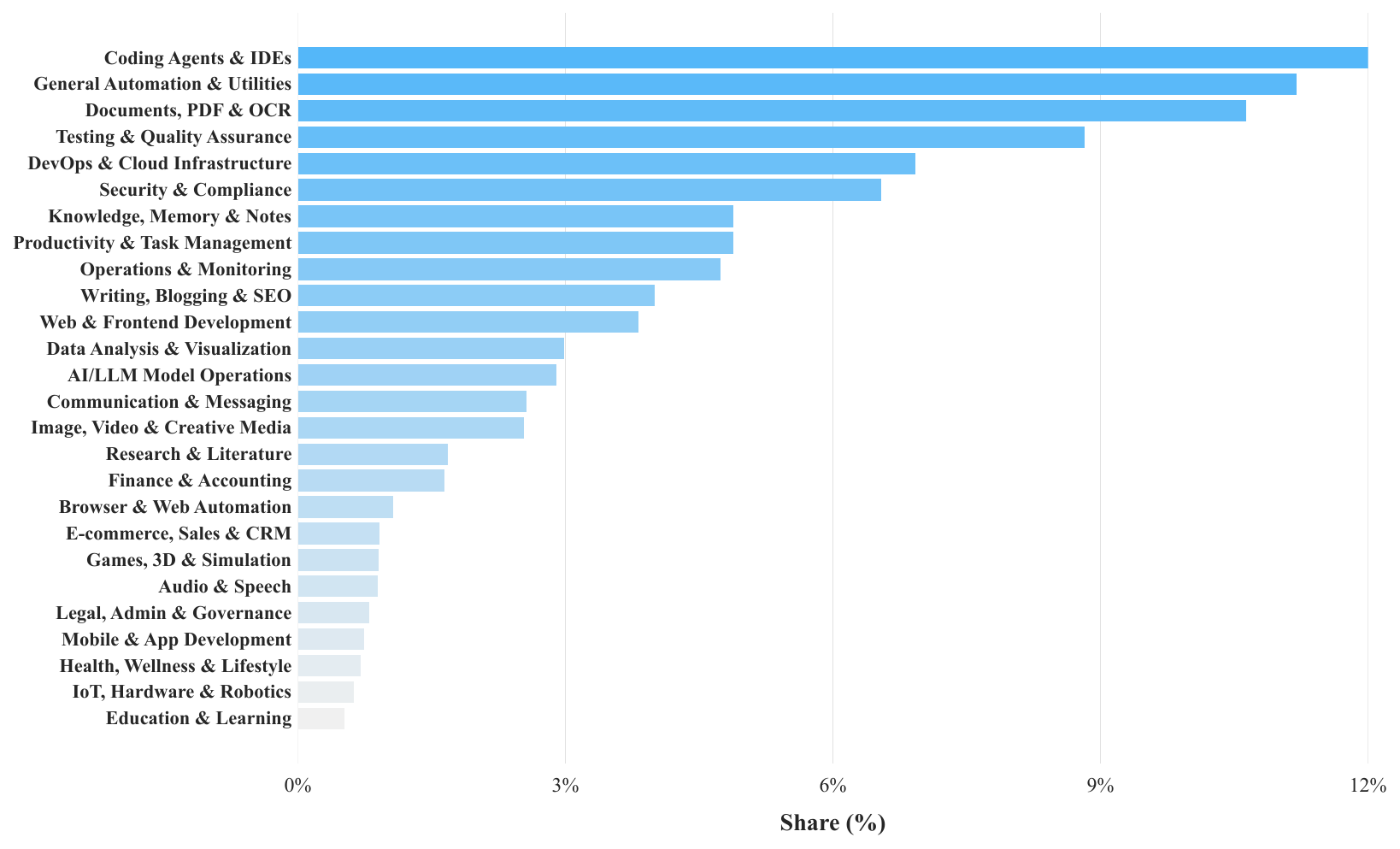}
    \caption{Skill category distribution.}
    \label{fig:skill_distribution}
\end{figure}
\begin{wrapfigure}{l}{0.45\textwidth}
\centering
\vspace{-1em}
\includegraphics[width=\linewidth]{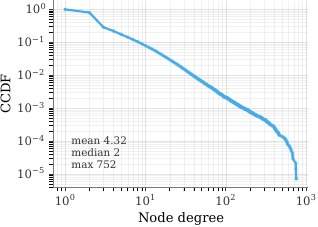}
\caption{Degree distribution of the skill graph.}
\label{fig:degree_distribution}
\end{wrapfigure}

The degree distribution (Figure~\ref{fig:degree_distribution}) is heavy-tailed: the median node degree is 2, while a small number of hub scenarios reach degree~752, corresponding to generic intermediate states compatible with many skills. This motivates the inverse-frequency sampling in \S~\ref{subsec:graph-guided-task-synthesis}. The connected component distribution (Table~\ref{tab:graph-stats}, right) is dominated by a single giant component of 118{,}806 nodes (85.6\%), confirming that the cross-skill alignment stage chains the majority of skills into a single traversable subgraph. Figure~\ref{fig:skill_distribution} shows that the graph covers both common domains, such as coding agents, document processing, DevOps, and security, as well as long-tail domains, such as audio/speech, 3D simulation and IoT/hardware workflows.

\section{Prompts}
\label{app:div_prompt}
The prompts used to extract skills and scenarios from trajectories, and to cluster embeddings to obtain unique instances, are provided below. We use DeepSeek Reasoner (v3.2) for extraction and Microsoft/Harrier-OSS-v1-27B for embedding.
\begin{promptbox}{Trajectory Analysis}
You are analyzing a terminal agent trajectory to extract its semantic structure.

TRAJECTORY:
{{{observation_action_sequence}}}

Your task: Segment this trajectory into a sequence of (scenario, skill) pairs, where:
- A SCENARIO is the semantic state of the environment at a decision point
  (e.g., "Python project with failing unit tests", "empty Git repository", 
  "server container with missing dependency").
- A SKILL is a coherent multi-step workflow the agent executes to transition 
  from one scenario to the next (e.g., "run pytest and capture output", 
  "install missing package via pip", "create a Dockerfile for the service").

Output a JSON array of pairs:
```json
[
  {
    "step_range": [start_index, end_index],
    "scenario": "<brief semantic description of the state before this skill>",
    "skill": "<brief description of the workflow executed in this range>"
  },
  ...
]
```

Keep scenario and skill descriptions under 15 words each.
Return ONLY the JSON array, no other text.
\end{promptbox}

\begin{promptbox}{Embedding Scenarios}
Instruct: Retrieve scenarios that describe the same real-world condition.
Query: {{{scenario}}}
\end{promptbox}

\begin{promptbox}{Embedding Skills}
Instruct: Retrieve skills that perform the same workflow.
Query: {{{skill}}}
\end{promptbox}

\end{document}